\documentclass[conference,a4paper]{IEEEtran}
\IEEEoverridecommandlockouts
\usepackage{cite}
\usepackage{amsmath,amssymb,amsfonts}
\usepackage{graphicx}
\usepackage{tikz}
\usetikzlibrary{arrows.meta, positioning, shapes.geometric, calc, backgrounds, fit}
\usepackage{booktabs}
\usepackage{enumitem}
\setlist[description]{style=nextline}
\usepackage{textcomp}
\usepackage{xcolor}
\usepackage{url}
\newif\ifblind
\blindfalse

\begin{document}

\title{Deterministic Conflict Resolution for Shared\\ Memory in Multi-Agent LLM Systems}

\ifblind
  \author{\IEEEauthorblockN{Anonymous Submission}
  \IEEEauthorblockA{Paper ID: XXX}}
\else
  \author{\IEEEauthorblockN{Maryem Jlassi}
  \IEEEauthorblockA{\textit{Computer Science} \\
  \textit{ESPRIT School of Engineering}\\
  Ariana, Tunisia\\
  Maryem.Jlassi@esprit.tn}
  \and
  \IEEEauthorblockN{Hadi Hasan}
  \IEEEauthorblockA{\textit{Electrical and Computer Engineering} \\
  \textit{American University of Beirut}\\
  Beirut, Lebanon \\
  hsh24@mail.aub.edu}
  \and
  \IEEEauthorblockN{Ali Chehab}
  \IEEEauthorblockA{\textit{Electrical and Computer Engineering} \\
  \textit{American University of Beirut}\\
  Beirut, Lebanon \\
  chehab@aub.edu.lb}
  }
\fi

\maketitle
\bstctlcite{IEEEexample:BSTcontrol}

\begin{abstract}
Multi-agent language model systems increasingly rely on shared, persistent memory to coordinate specialized agents, yet existing memory mechanisms provide limited support for resolving contradictory writes in a deterministic, auditable manner. Conventional approaches such as last-write-wins, summarization, and model-based judging may overwrite valid information, depend on non-deterministic decisions, or allow repeated unsupported claims to gain legitimacy. This paper presents the Conflict Resolution Tracer (CRT), a deterministic, zero-LLM middleware for conflict resolution and concurrency control in shared agent memory. CRT enforces a fixed write pipeline comprising provenance validation, oscillation detection, per-path locking, evidence-based arbitration, and immutable lineage tracking. Conflicting claims are scored deterministically from recency, middleware-verified evidence confidence, and domain-specific historical trust, with agent-reported confidence excluded from arbitration. When competing claims are insufficiently distinguishable, CRT abstains and preserves both alternatives rather than forcing an arbitrary decision. We evaluate CRT on three heterogeneous benchmarks spanning real-world rumor verification, multi-provider question answering, and synthetic multi-agent memory. Compared with last-write-wins, CRT improves selective accuracy on every track, significantly so only on the synthetic MSM benchmark, and lowers the incorrect-overwrite rate by 67.8\% on QACC and 45.3\% on MSM while deciding fewer cases. These gains come at the cost of lower coverage, an explicit precision--coverage trade-off introduced by abstention. The results show that deterministic middleware can deliver reproducible, auditable conflict resolution while preserving disputed information and reducing silent state corruption.
\end{abstract}


\section{Introduction}
\label{sec:introduction}

Language models are now deployed as teams of specialized agents (e.g., triage, policy lookup, drafting) that share a long-lived memory~\cite{hong2024metagpt}, increasingly in high-stakes domains such as healthcare~\cite{hasan2026toward}. The arrangement only works if that memory stays coherent, and nothing guarantees it will, because each agent is a statistical predictor with no self-consistency guarantee~\cite{ji2023survey}. Existing solutions treat coherence as a problem of scale. Larger contexts, summarization, and retrieval~\cite{hasan2025kag} each stretch how much the agents can see, while live contradiction resolution is left to ad-hoc heuristics~\cite{zhang2025survey}. The interoperability protocols standardize transport, discovery, and tool invocation and push state reconciliation down to the application layer~\cite{mcp2025spec}.

Three failures follow. The last write in a race silently overwrites a correct one, and early facts in a long context become invisible because of where attention lands~\cite{liu2024lost}. A claim repeated often enough starts to pass for one backed by evidence. Each of these comes from a missing reconciliation rule, and a stronger model would leave all three in place.

We introduce the Conflict Resolution Tracer (CRT), a deterministic middleware that sits between the agents and storage. Every write follows one fixed, inspectable path that can be replayed to the same outcome, and a close call is kept as two alternatives.

\textbf{Research questions.} We ask whether conflicting claims can be adjudicated reproducibly (RQ1), concurrent writes sequenced without agent coordination (RQ2), legitimacy by repetition prevented (RQ3), and provenance and trust modeled so decisions stay auditable as reliability accumulates (RQ4).

\textbf{Contributions.}
\begin{enumerate}[leftmargin=*,itemsep=1pt,topsep=2pt]
  \item A deterministic write pipeline with no LLM in the loop, in which every conflict decision can be replayed and explained from its component scores (Sec.~\ref{sec:methodology}).
  \item An arbitration score $\Psi$ that resists manipulation by construction. Agent-reported confidence is stored for audit and never consumed, and trust is isolated per domain.
  \item Abstention as a first-class outcome. When the score gap falls below a threshold, both claims are preserved.
  \item An evaluation across three tracks that quantifies the resulting precision/coverage trade-off against last-write-wins and against weight ablations (Sec.~\ref{sec:experiment}).
\end{enumerate}

\section{Related Work}
\label{sec:related}

\subsection{Multi-Agent LLM Memory and the Reconciliation Gap}
Recent frameworks organize LLMs into role-specialized teams that share a workspace~\cite{hong2024metagpt}, and their members keep memory streams they periodically reflect on~\cite{park2023generative}. Surveys of agent memory catalogue read, write, and reflect mechanisms~\cite{zhang2025survey}. In every surveyed design the memory is append-only or summarization-driven, and none of them says what happens when two agents assert incompatible values for the same key. Interoperability standards delegate that question to the application layer~\cite{mcp2025spec}. The gap matters because LLMs contradict both themselves and their sources. Hallucination is well documented~\cite{ji2023survey}, attention placement hides mid-context facts~\cite{liu2024lost}, and models are neither consistently faithful to retrieved evidence nor consistently stubborn against it~\cite{xie2024adaptive}. The usual mitigation is an LLM judge that scores the competing candidates~\cite{zheng2023judging}. Such a judge shows positional and verbosity bias, shifts with prompt order~\cite{wang2024large}, and relies on verbalized confidence that is self-reported and gameable~\cite{lin2022teaching}. A learned judge is also exposed to adaptive adversarial inputs~\cite{hasan2026defending}. Since its decisions cannot be replayed, it also cannot furnish the audit trail a memory of record requires. CRT supplies the missing policy layer and removes the model from arbitration entirely.

\subsection{Truth Discovery, Provenance, Trust and Abstention}
Reconciling claims from sources of unequal quality is a long-standing data-integration problem. Truth discovery jointly estimates source reliability and fact veracity~\cite{yin2008truth} and corrects for copying between sources~\cite{dong2009integrating}, but both are global fixed-point computations, and a live shared memory needs to decide a single write synchronously. Provenance research supplies why-, how-, and where-lineage~\cite{herschel2017survey}, which CRT instantiates as a hash-linked chain. Reputation systems supply longitudinal source quality~\cite{josang2007survey} along with the identity attacks it invites~\cite{douceur2002sybil}, which is why CRT isolates trust strictly per domain. Symbolic truth maintenance formalizes retraction under contradiction~\cite{doyle1979truth}, though it presumes logical sources where ours are statistical. The reject option, which trades coverage against error~\cite{chow1970optimum} and was later formalized as selective classification~\cite{geifman2017selective}, is the classical form of withholding a decision and pays most where errors are costly, as in clinical screening~\cite{hasan2026end}. Neural confidence is poorly calibrated~\cite{guo2017calibration} and accuracy varies across architectures on a fixed task~\cite{hasan2026comprehensive}, so CRT scores evidence the middleware has verified, ignores model-reported certainty, and reports Wilson intervals~\cite{wilson1927probable}.

\subsection{Concurrency Control and Replica Conflict Resolution}
Databases solved write races with locking and validation decades ago~\cite{bernstein1981concurrency}. Weakly consistent stores must also merge divergent replicas. Bayou attaches application-supplied merge procedures~\cite{terry1995managing}, Dynamo exposes sibling versions and in practice resolves by last-write-wins~\cite{decandia2007dynamo}, and CRDTs converge by algebraic construction~\cite{shapiro2011conflict}. Peer-to-peer energy markets pose a related coordination problem~\cite{chedid2026model}. Convergence, though, says nothing about correctness, since last-write-wins follows whichever write carries the larger timestamp regardless of evidence, and a timestamp order encodes causality alone~\cite{lamport1978time}. Agreement protocols likewise settle which write is ordered first and leave open which claim is true. CRT keeps the systems discipline of per-key mutual exclusion and durable lineage, replaces timestamp dominance with an evidence-weighted score, and treats a near-tie as an abstention in its own right.

\section{Methodology}
\label{sec:methodology}

\subsection{Problem Formulation}
Let $\mathcal{A} = \{a_1, a_2, \ldots, a_n\}$ be a set of agents sharing a common memory store $\mathcal{M}$. The store is a key-value map from \textit{memory paths} $p \in \mathcal{P}$ to \textit{stamped claim packets} $u \in \mathcal{U}$. A claim packet $u$ carries a payload, a timestamp, evidence, and a self-reported confidence, the last of which is stored for audit and never used in arbitration.

When agent $a_i$ writes to path $p$ and $\mathcal{M}(p) = \emptyset$, the claim is committed directly. When $\mathcal{M}(p) = u_{\mathit{ex}}$, the incoming claim $u_{\mathit{in}}$ conflicts with it, and arbitration must decide which of the two to commit, or whether the conflict cannot be settled at all.

An arbitration rule that does this job needs five properties: \textit{manipulation resistance}, \textit{calibrated abstention}, \textit{longitudinal trust} with decay, \textit{determinism} (identical inputs give identical outcomes), and \textit{auditability} by component scores.

\subsection{Architecture Overview}
CRT has three logical layers (see Fig.~\ref{fig:pipeline}), and the split keeps probabilistic agent reasoning apart from deterministic state management. \textit{Layer 1 (Agent Layer):} LLM-enabled agents reach shared memory only through the HTTP Client SDK and submit a six-field raw Universal Memory Format (UMF) packet. They cannot supply provenance or authority metadata. \textit{Layer 2 (CRT Core):} the deterministic core runs the four-stage pipeline with no LLM anywhere in it. \textit{Layer 3 (Persistent Storage):} holds active, pending, and archived packets together with lineage, trust, and provenance records.

\subsection{Universal Memory Format}
\subsubsection{UMF Schema and Stamping}
CRT separates what an agent provides from what the middleware verifies. A write begins as a lean 6-field \textbf{Universal Memory Format (UMF)} packet whose schema is configured with \texttt{extra="forbid"}, so an agent cannot spoof authority or confidence. The middleware then stamps six immutable fields: \texttt{provenance\_id}, \texttt{ingested\_at}, \texttt{content\_hash}, \texttt{source\_type/id}, \texttt{authority\_score} and \texttt{verified\_confidence}, all computed in Stage~1 and frozen before Stage~3.

\subsection{Four-Stage Pipeline}
\begin{figure*}[t]
\centering
\includegraphics[width=0.65\textwidth]{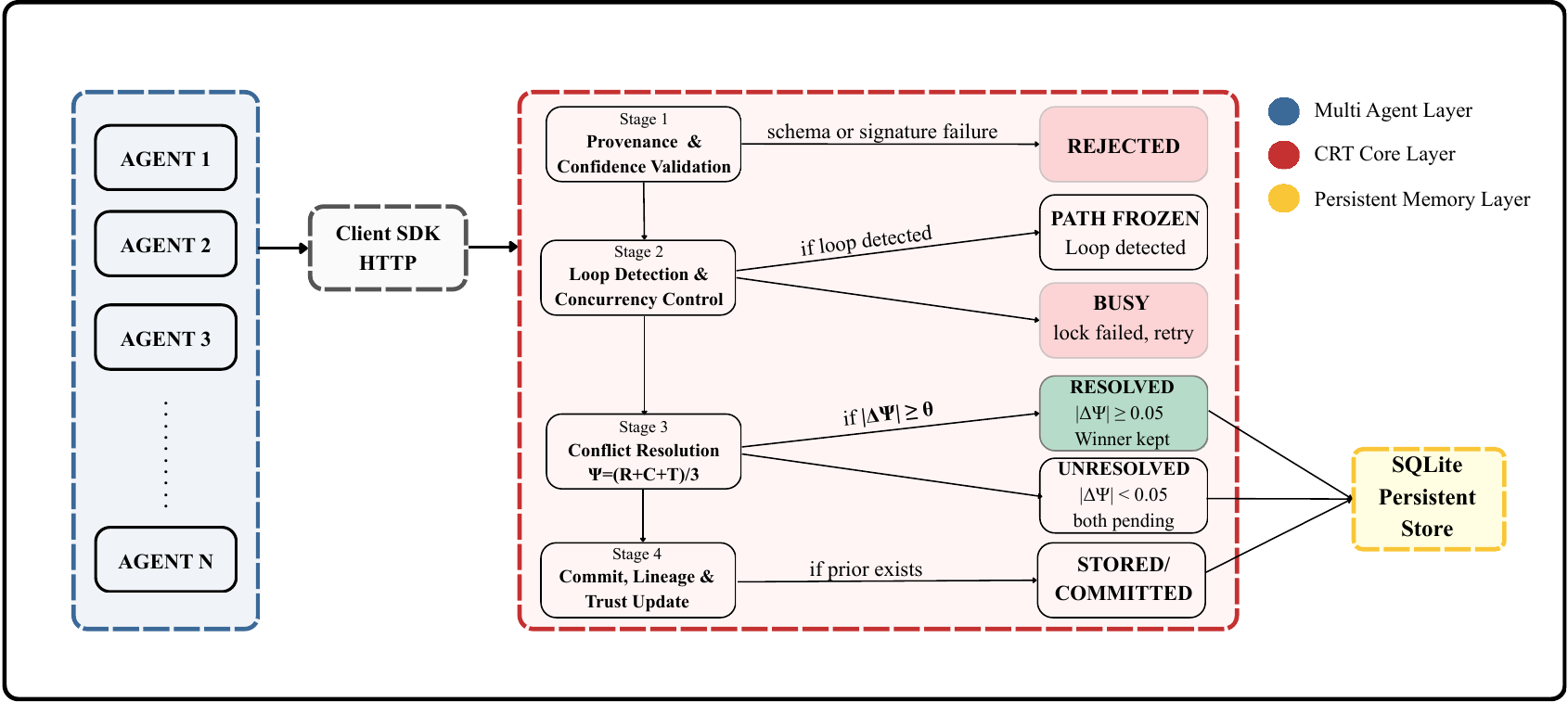}
\caption{CRT write pipeline. Agent claims enter through the Client SDK and traverse four deterministic stages. Invalid claims are rejected, oscillating paths are frozen, and lock contention produces a retryable busy outcome. Conflicts are either resolved using $\Psi$ or retained as an active claim plus a pending alternative.}
\label{fig:pipeline}
\end{figure*}

Every incoming claim passes through the four stages shown in Fig.~\ref{fig:pipeline}. Any stage can reject or freeze the claim on its own, and only a claim that clears all four reaches a durable commit.

\textit{Stage 1 (Provenance Validation):}
Each raw claim is validated against a strict schema (\texttt{extra="forbid"}). Unknown fields, and any middleware field an agent tries to supply itself, cause a fail-closed rejection. Verified confidence is computed from the highest-ranked evidence record as $C = \alpha(\mathrm{type}) \cdot \mathrm{relevance}$, where $\alpha$ comes from a provider-blind authority hierarchy: 1.00 for gateway-attested user input, 0.90 for a database record, 0.85 for tool output such as a sensor reading~\cite{hasan2022development}, 0.75 for a document reference, 0.30 for a bare agent claim, and 0.10 for unsigned or unattested evidence. For elevated evidence CRT verifies a canonical Ed25519 binding. An invalid, expired, or replayed signature causes rejection, and an unsigned claim drops to the lowest authority tier.

\textit{Stage 2 (Loop Detection and Locking):}
A per-path sliding window tracks write rate, value oscillation, and which agents are involved. If oscillation crosses a threshold, the request exits with \textsc{Loop Frozen} before any lock is taken, and this one check is what blocks legitimacy by repetition. Otherwise the pipeline tries to acquire a token-based per-path lock, which is released once compare-and-commit completes or the write aborts. If another writer holds the lock, the request exits with \textsc{Lock Failed} and may be retried.

\textit{Stage 3 (Conflict Resolution):}
When an incoming claim contradicts the existing record, the Conflict Resolution Engine computes a scalar score $\Psi \in [0,1]$:
\begin{equation}
  \Psi(u) = w_R \cdot R(u) + w_C \cdot C(u) + w_T \cdot T(u)
  \label{eq:psi}
\end{equation}
with weights $(w_R, w_C, w_T)$ summing to 1 and defaulting to $(1/3, 1/3, 1/3)$ (see Sec.~\ref{sec:discussion}). The three components are:

\begin{itemize}[leftmargin=*]
  \item $R(u)$: \textbf{Recency}, an exponential decay from the claim timestamp to the shared reference time $t_{\mathit{ref}}$,
    \[
      R(u) = e^{-\lambda (t_{\mathit{ref}} - u.\mathit{timestamp})},
    \]
    where the decay constant $\lambda = \ln 2 / h_R$ is set by a deployment half-life $h_R$ (default $24$\,h; per-track values in Sec.~\ref{sec:experiment}).

  \item $C(u)$: \textbf{Verified Confidence}, the evidence score the middleware computed in Stage 1. The confidence the agent reported about itself plays no part in it.

  \item $T(u)$: \textbf{Historical Trust}, the agent's domain-specific reliability score from the trust manager, strictly isolated per domain, with Wilson 95\% intervals and a 30-day decay toward neutral.
\end{itemize}

\textbf{Decision rule.} Let $\delta = |\Psi(u_{\mathit{in}}) - \Psi(u_{\mathit{ex}})|$ and let $\tau$ be the uncertainty threshold (default $0.05$). Scores are quantized to 12 decimal places before comparison so that floating-point ties resolve the same way on every platform. The engine applies:
\[
  \text{outcome} =
  \begin{cases}
    u_{\mathit{in}} \text{ wins} & \Psi(u_{\mathit{in}}) > \Psi(u_{\mathit{ex}}),\ \delta \geq \tau \\[2pt]
    u_{\mathit{ex}} \text{ wins} & \Psi(u_{\mathit{ex}}) \geq \Psi(u_{\mathit{in}}),\ \delta \geq \tau \\[2pt]
    \text{unresolved} & \delta < \tau
  \end{cases}
\]
When the outcome is \textsc{Unresolved}, both claims are kept.

\textit{Stage 4 (Commit and Lineage):}
The resulting state is persisted to the SQLite store, where a winner is committed as \texttt{active}. A loser is archived, never deleted. When the outcome is unresolved, the existing record stays active and the incoming claim is stored as a pending conflict. Each record gets a lineage node that binds a SHA-256 content hash to its parent provenance identifiers, so a stored hash that disagrees with the recomputed digest exposes an inconsistent record.

\subsection{Trust Modeling}
Trust $T$ is an observed statistic kept per agent and per domain. It changes only through externally adjudicated outcomes, is penalized for small samples, decays toward uncertainty, and never crosses domains. The raw score is $T_{\text{raw}}(a,d)=\text{correct}(a,d)/\text{total}(a,d)$, with a neutral cold start $T=0.5$ when $\text{total}=0$. For high-stakes gates CRT uses the Wilson 95\% lower bound as \textit{conservative trust}:
\begin{equation}
T_{\mathrm{Wilson}} = \frac{\hat p + \frac{z^2}{2n} - z\sqrt{\frac{\hat p(1-\hat p)}{n} + \frac{z^2}{4n^2}}}{1+\frac{z^2}{n}},
\label{eq:wilson}
\end{equation}
where $\hat p = \text{correct}/\text{total}$, $n = \text{total}$, and $z=1.96$~\cite{wilson1927probable}. Trust decays toward the neutral prior with a 30-day half-life,
$T_{\text{now}} = 0.5 + (T_{\text{last}}-0.5)\,e^{-\gamma\Delta t}$, $\gamma=\ln 2/(30\times86400)$,
so an agent that was reliable months ago but has since gone quiet drifts back toward $0.5$. This decay is separate from recency $R$, which acts on a different signal at a different timescale. Domain isolation admits no exception: an agent trusted in \texttt{coding} starts at neutral cold-start trust the first time it acts in \texttt{medical}, with no silent fallback to a global average, which stops a reputation earned in one area from overriding a specialist in another and caps what a compromised or Sybil identity can accumulate~\cite{douceur2002sybil}.

\section{Experiments and Results}
\label{sec:experiment}

\subsection{Experimental Setup}
We evaluate CRT on three tracks chosen to differ in character. Each source is stamped once with a normalized timestamp, and the frozen claim set is reused unchanged across every resolver policy. Provider identity is checked fail-closed, so a mismatch is recorded as a failure and never silently substituted. Proportions carry Wilson 95\% intervals computed from exact counts~\cite{wilson1927probable}.

\textit{PHEME:} Real-world rumor verification~\cite{zubiaga2016analysing} over 9 events, 6,425 threads, 105,354 tweets, and 1,950 evaluable instances. The gold label is event- and thread-level veracity from the PHEME annotation. Recency comes from tweet timestamps with a 7-day half-life, confidence from a lexicon-based stance extractor, and trust from training events only, with no cross-contamination.

\textit{QACC:} Multi-provider question answering over 500 questions, 4,823 sources, and 495 evaluable cases. The gold label is the answer adjudicated against frozen web evidence. Recency and trust are held at a neutral 0.5, which leaves evidence-derived confidence $C$ as the only signal separating competing claims.

\textit{MSM:} A synthetic structural benchmark with 4 seeds, 192 personas, 3,456 units, and 3,337 claim-bearing units. The gold label is a deterministic structural readout for each persona--question pair. Recency $R$ comes from synthetic timestamps with a 30-day half-life, confidence is $C = \alpha(\text{source type}) \cdot \text{coverage}$, and trust $T$ is per-source per-question reliability estimated from training units.

Determinism is the central claim, so every reported number can be regenerated from the frozen claim sets and a fixed resolver configuration. Implementation, claim sets, and analysis scripts will be released on publication.

\subsection{Baselines and Metrics}
\begin{table*}[!t]
\centering
\caption{Per-policy results on all three tracks. Sel = selective accuracy (correct/decided), Cov = coverage (decided/all), OW = incorrect overwrites as a count and as a percentage of decided cases. Strict accuracy is the product Sel $\times$ Cov. Compare policies by OW \emph{rate}, since the raw count depends on how many cases each policy decided, and recency-only's zero on MSM rests on just 192 decisions. PHEME OW cannot be evaluated because its gold labels are thread-level veracity.}
\label{tab:ablation}
\scriptsize
\begin{tabular}{lccc ccc ccc}
\hline
& \multicolumn{3}{c}{PHEME} & \multicolumn{3}{c}{QACC} & \multicolumn{3}{c}{MSM}\\
Method & Sel & Cov & OW & Sel & Cov & OW $n$ (\%) & Sel & Cov & OW $n$ (\%)\\
\hline
full & 17.66 & 73.74 & --- & 42.16 & 20.61 & 4 (3.9) & 69.17 & 57.73 & 254 (12.7)\\
evidence-only & 13.59 & 100.0 & --- & 41.75 & 20.81 & 4 (3.9) & 57.74 & 83.74 & 701 (24.2)\\
trust-only & 15.54 & 100.0 & --- & 42.35 & 17.17 & 0 (0.0) & 63.97 & 53.96 & 248 (13.3)\\
recency-only & 16.00 & 100.0 & --- & 42.35 & 17.17 & 0 (0.0) & 45.83 & 5.56 & 0 (0.0)\\
fixed-neutral & 13.59 & 100.0 & --- & 42.16 & 20.61 & 4 (3.9) & 63.20 & 32.55 & 174 (15.5)\\
last-write & 16.00 & 100.0 & --- & 38.37 & 89.49 & 54 (12.2) & 58.83 & 96.56 & 777 (23.3)\\
\hline
\end{tabular}
\end{table*}
We compare the full CRT rule $(w_R,w_C,w_T)=(1/3,1/3,1/3)$ against evidence-only $(0,1,0)$, trust-only $(0,0,1)$, recency-only $(1,0,0)$, fixed-neutral-trust ($T=0.5$), and last-write-wins (LWW). All policies receive identical stamped claims, and the weight-ablation policies differ from the full rule only in resolver weights.

We report four quantities. \textit{Coverage} is the fraction of instances resolved. \textit{Strict accuracy} counts every abstention as an error, so it is the share of correctly resolved cases over all cases. \textit{Selective accuracy} is the share of resolved cases that are correct. An \textit{incorrect overwrite} (OW) occurs when the gold value is present among the claims, CRT decides, and the winner it picks is wrong. Strict and selective accuracy sit at the two ends of the reject-option trade-off~\cite{chow1970optimum}. Abstaining can only lower strict accuracy, so it pays off only when it buys precision and spares correct state.

\textit{Component discrimination} measures how often a single component decides the outcome, with a different probe per track. On PHEME we use a \emph{removal} probe: zero one component's weight and record whether the decision flips. On MSM, where the synthetic design lets components vary independently, we use an \emph{argmax} probe: find which claim wins on that component alone with the others at a neutral 0.5, and match it against the full-rule winner. The two probes answer different questions, so their values should not be compared across tracks.

\textit{Provider and Transport Failures:}
For QACC each source was assigned to one provider by round-robin (1,612 OpenAI, 1,603 Ollama, 1,608 Groq). A separate 30-source sub-experiment queried all three providers and yielded the 7 cases on which agreement could be measured. Only successfully parsed outputs entered the frozen set.
A Groq transport failure is an HTTP 429 rate limit or an unparseable payload after a single attempt, and 1,357 of the 1,608 Groq requests (84.4\%) failed this way. A failed request produces no claim and never enters the frozen memory set, so the Groq portion of the set is simply smaller, and we draw no conclusion about provider quality from it. Provider blindness held throughout: across the 4,823 assigned sources there were 0 identity mismatches.

\subsection{Results}
\begin{figure*}[!t]
\centering
\begin{minipage}[t]{0.47\textwidth}
\centering
\includegraphics[width=\linewidth]{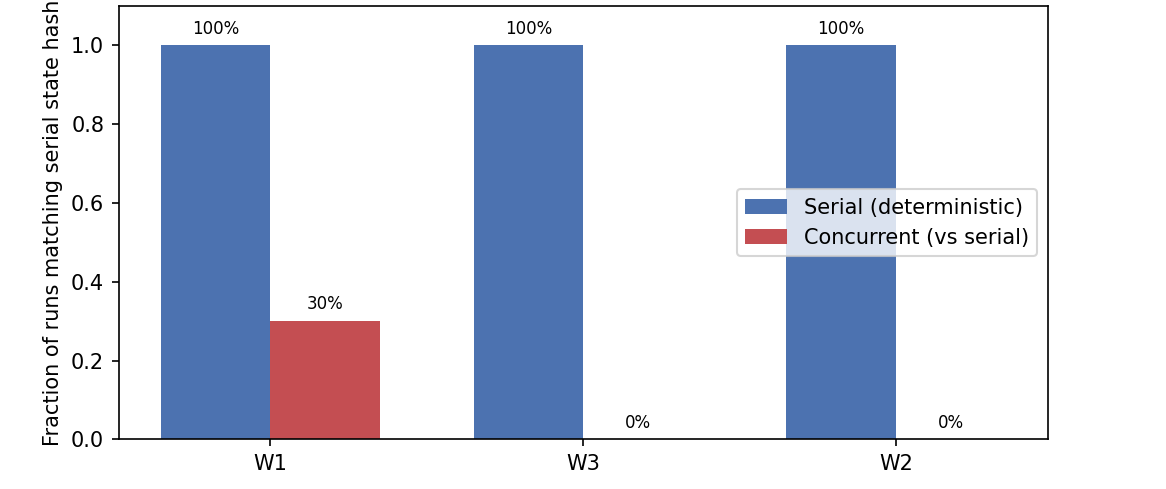}
\caption{Determinism under serial and concurrent submission across three workloads. The state hash
combines record content with the \texttt{active}/\texttt{pending} label. Serial replay reproduces the
same hash in every run. Concurrent submission reproduces it in 30\% of runs on W1 and in none on W2
or W3. All operations are preserved in every workload (30/30, 12/12 and 20/20), so a hash mismatch
reflects variation in labels alone, with no data lost.}
\label{fig:determinism}
\end{minipage}\hfill
\begin{minipage}[t]{0.428\textwidth}
\centering
\includegraphics[width=\linewidth]{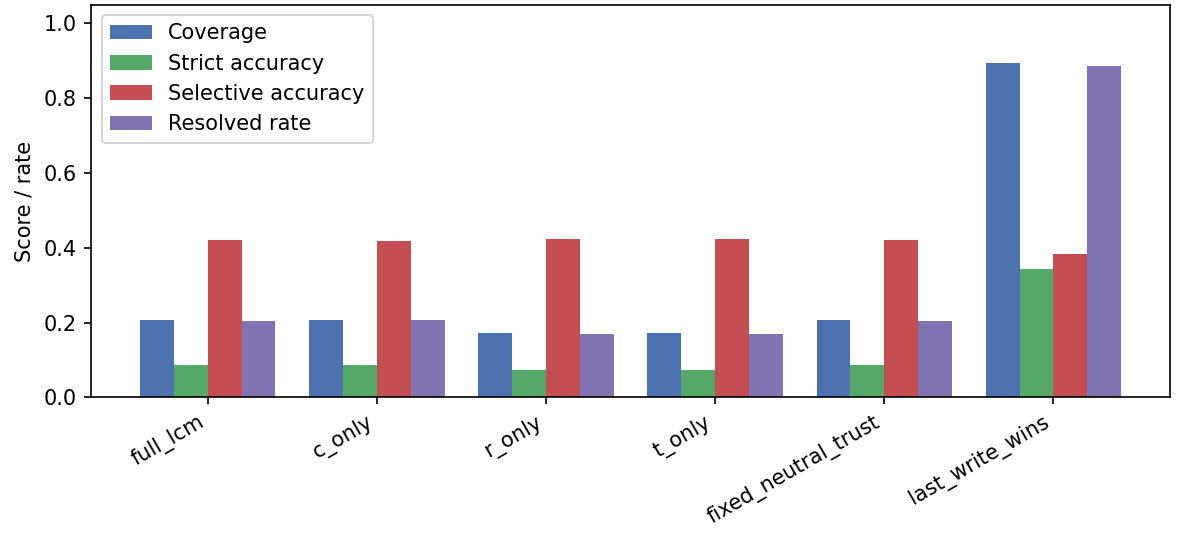}
\caption{QACC policy comparison over 495 evaluable cases. Coverage, strict accuracy, selective
accuracy and resolved rate for the full rule, the four weight ablations and last-write-wins. The
full rule and the ablations cluster at high selective accuracy and low coverage. Last-write-wins
buys coverage and strict accuracy at a visibly lower selective accuracy. Trust-only and recency-only
coincide because $R$ and $T$ are neutral on this track by construction.}
\label{fig:qacc}
\end{minipage}
\end{figure*}
Table~\ref{tab:ablation} reports every policy on every track. The pattern on all three is a precision/coverage trade. CRT decides fewer cases than LWW and is more often right when it does decide. Selective accuracy rises from 16.00\% to 17.66\% on PHEME, from 38.37\% to 42.16\% on QACC (Fig.~\ref{fig:qacc}), and from 58.83\% to 69.17\% on MSM. Under a two-proportion $z$-test only the MSM gain is significant ($z=7.56$, $p<10^{-4}$). The PHEME and QACC gains sit inside sampling noise ($p=0.20$ and $p=0.48$), so we claim no precision improvement on those two tracks. Strict accuracy, which charges every abstention as an error, moves the other way: 39.93\% [38.30, 41.56] for the full rule against 56.80\% [55.13, 58.43] for LWW on MSM (Wilson 95\%). That is the arithmetic of deciding 57.73\% of cases where LWW decides 96.56\%.

Incorrect overwrites, the quantity that matters for safety, fall on QACC from 54 under LWW to 4 under the full rule, a 92.6\% drop in count and a 67.8\% drop once normalized by the number of decisions (12.19\% to 3.92\% of decided cases). On MSM they fall from 777 to 254, a 67.3\% drop in count and 45.3\% in rate (23.28\% to 12.73\% of decided cases). Both rate reductions are significant ($p=0.015$ on QACC, $p<10^{-4}$ on MSM), unlike the selective-accuracy gains, so abstention lands disproportionately on the cases LWW gets wrong.

\textit{Determinism and concurrency.} Fig.~\ref{fig:determinism} isolates the reproducibility claim. Serial replay reproduces an identical state hash in every run on all three workloads, which is the property the arbitration algebra is built to guarantee. Concurrent submission of the same operations reproduces it in 30\% of runs on W1 and in none on W2 or W3. The divergence is confined to which claim carries the \texttt{active} label. No update is lost. Concurrency costs only the reproducibility of the label assignment.

On PHEME the lexicon stance extractor is the binding constraint. Verified confidence dominates, recency and trust contribute little on noisy short text, and every single-component policy decides all 1,950 instances while the full rule abstains on 26\%. Averaging three components shrinks the score gap by roughly a factor of three whenever two of them barely separate the candidates, which pushes $\delta$ below $\tau$. On QACC, $R$ and $T$ are neutral by construction, so evidence is the only differentiating signal. Trust-only coincides with recency-only, full coincides with fixed-neutral-trust, and the resulting ties explain the low coverage, with the full rule deciding 102 cases against 103 for evidence-only. The OpenAI win share cannot be distinguished from its share of contributed sources. On MSM, where evidence types and source reliabilities are structurally separable, the full rule draws on all three components. Recency-only is degenerate at 5.56\% coverage because synthetic timestamps tie often, and dropping trust costs the most, with evidence-only nearly doubling the overwrite rate to 24.2\%. Judged by overwrite rate, the one quantity comparable across policies, the full rule leads every non-degenerate policy on both evaluable tracks (3.9\% on QACC, 12.7\% on MSM).

Component discrimination (Sec.~IV-B) shows which signal drives decisions. On PHEME the removal probe flips the decision for evidence in 30.41\% of cases, for trust in 10.36\%, and for recency in 7.44\%, consistent with the lexicon extractor being the binding constraint. On MSM the argmax probe agrees with the full-rule winner for trust in 66.2\% of decided cases, for recency in 58.8\%, and for evidence in 57.8\%. No single component reproduces the full rule, so the equal weights sit on a flat plateau with no sharp optimum in sight.

\section{Discussion}
\label{sec:discussion}

\textit{Design and abstention.} Constant weights and a canonical $\Psi$ make identical inputs yield identical outcomes, and everything else in the design rests on that. The most consequential choice is to ignore self-reported confidence. Doing so removes inflation attacks~\cite{lin2022teaching}, at the cost of a flat per-type authority hierarchy that does not model quality within a type and that a valid key-holder can still game by misrepresenting content type. Abstention is equally deliberate, and the results back it, since overwrite rates fall significantly on both measurable tracks. Trust is learned only from external verification and isolated per domain, which blocks repetition attacks and cross-domain laundering. Cold-start agents receive neutral trust, and CRT cannot repair a weak evidence extractor.

\textit{Weight selection.} The equal weights are a static default, and the component probes show that the informative signal changes with the domain (evidence on PHEME, trust on MSM). The adjudicated outcomes that update trust could also fit per-domain weights and $\tau$ by minimizing an expected cost $c_{\mathrm{ow}}P_{\mathrm{ow}} + c_{\mathrm{ab}}P_{\mathrm{ab}}$ over the incorrect-overwrite and abstention rates, as in the cost-optimal reject rule~\cite{chow1970optimum}. A clinical memory, where a wrong commit far outweighs a deferral~\cite{hasan2026toward}, calls for a large $c_{\mathrm{ow}}/c_{\mathrm{ab}}$, whereas time-critical advice tolerates less deferral. Weights could also adapt online via reinforcement learning~\cite{hasan2026defending}. To stay deterministic, learned weights must be frozen into versioned epochs recorded in the lineage, so each decision replays under the weights then in force, and updates bounded, since adaptation hands a lever to whoever influences adjudication.

\textit{Overhead and limitations.} Excluding LLM generation, mean committed-write latency is 499.6\,ms on the compatible workload (W2) and 841.3\,ms on the conflicting one (W3), against 892.7\,ms and 1,628.6\,ms under a fully serial store, a 1.8 to 1.9$\times$ improvement, and per-path locking confines serialization to same-path writes. Strict accuracy stays below last-write-wins on all three tracks, so a deployment that must answer every query and can tolerate silent corruption would prefer LWW, and the sweep does not establish $(1/3,1/3,1/3)$ as optimal. On real data, evidence extraction is the binding constraint. The 84.4\% Groq failure rate rules out any provider-quality comparison. Concurrent distinct-path submission can surface spurious pending conflicts (Fig.~\ref{fig:determinism}), a process-local race that loses no data but needs fixing before multi-writer deployment, and distributed concurrency control~\cite{bernstein1981concurrency} remains future work.

\section{Conclusion}
We presented the Conflict Resolution Tracer (CRT), a deterministic middleware with no LLM in the loop that enforces memory coherence in multi-agent systems. By replacing probabilistic judging with a fixed algebra, CRT turns shared memory into a controlled, auditable knowledge store, and it gives a principled way to abstain and preserve history when the available evidence cannot separate two conflicting claims.

Across three benchmark tracks CRT behaves as a precision-oriented resolver. It significantly reduces the incorrect-overwrite rate on both tracks where gold values are recoverable (68\% on QACC and 45\% on MSM among decided cases) and improves selective accuracy on every track, significantly on the synthetic one. It decides fewer cases and so scores lower on strict accuracy. For a memory of record, where a silently corrupted fact is the expensive error and a deferred one is cheap, that is the intended trade. Future work will strengthen evidence extraction for unstructured inputs, learn per-domain weights and $\tau$ from explicit decision costs, and extend path-level concurrency control to distributed storage.

\bibliographystyle{IEEEtran}
\bibliography{references}

\end{document}